%% file: ms.tex
\documentclass{article}

\PassOptionsToPackage{numbers}{natbib}
\usepackage{amsmath}
\usepackage[preprint]{nips_2018}
\usepackage[utf8]{inputenc}
\usepackage[T1]{fontenc}
\usepackage{booktabs}
\usepackage{graphicx}
\usepackage{wrapfig}
\usepackage{numprint}
\usepackage{caption}
\title{Holographic Neural Architectures}

\author{Tariq Daouda\\Institute for Research in Immunology and Cancer\\Department of biochemistry\\Université de Montréal
\\\texttt{tariq.daouda@umontreal.ca}
			\And Jeremie Zumer\\Institute for Research in Immunology and Cancer \\Department of Computer Science and Operations Research\\Université de Montréal
            \\\texttt{jeremie.zumer@umontreal.ca}\\
			\And Claude Perreault\\Institute for Research in Immunology and Cancer \\Department of Medicine\\Université de Montréal\\\texttt{claude.perreault@umontreal.ca}\\
			\And Sébastien Lemieux\\Institute for Research in Immunology and Cancer \\Department of Computer Science and Operations Research\\Université de Montréal\\\texttt{s.lemieux@umontreal.ca}
			}

\begin{document}
\maketitle
\let\clearpage\relax
\include{abstract}

\include{intro}
\include{arch}
\include{hgn}
\include{bio}
\include{conclusion}
\bibliography{ms}
\end{document}

%% file: abstract.tex
\begin{abstract}
Representation learning is at the heart of what makes deep learning effective. In this work, we introduce a new framework for representation learning that we call ``Holographic Neural Architectures'' (HNAs). In the same way that an observer can experience the 3D structure of a holographed object by looking at its hologram from several angles, HNAs derive Holographic Representations from the training set. These representations can then be explored by moving along a continuous bounded single dimension. We show that HNAs can be used to make generative networks, state-of-the-art regression models and that they are inherently highly resistant to noise. Finally, we argue that because of their denoising abilities and their capacity to generalize well from very few examples, models based upon HNAs are particularly well suited for biological applications where training examples are rare or noisy.
\end{abstract}

%% file: intro.tex
\section{Introduction}
The success of deep learning algorithms during the last decade can be attributed to their ability to learn relevant latent representations \citep{bengio2013representation}. However, despite these recent advances, generalization from few and potentially noisy examples remains an open problem. Most deep learning architectures have many more parameters than the number of examples they are trained upon and are therefore particularly sensitive to noise and overfitting on small datasets. These limitations can make deep learning algorithms hard to apply to real life situations where data is scarce and noisy. These problems could in theory be overcome if we were able to effectively learn relevant, smooth, latent representations from few examples and despite the presence of noise. 

Manifold learning is an emergent property of some deep learning algorithms. A class of algorithms that aims at deriving latent, lower-dimensional manifolds from the training set are Autoencoders (AEs) with bottleneck layers  \citep{reprlearning}. Since these networks reconstruct their input using fewer units than the input dimensionality, it is assumed that the bottleneck representation `summarises' the input. More recently, the application of variational approaches to auto-encoders have led to the introduction of Variational Autoencoders (VAEs) \citep{vae}. VAEs address the problem of manifold learning by constraining the parameter distribution to a simpler distribution family than the true distribution. This approach forces the latent space to be dense, enabling straightforward sampling for generation.

In our approach, manifold learning is also a consequence of the network's training. We force the network to generalize by placing an extremely severe bottleneck over the information received from training examples. Here, we are projecting all training example into a single bounded dimension. As with VAEs, we also combine the input information with an optimized prior. However, we treat the prior as a separate input to the network.
Because the network has very little information from the training examples, it must complement it with an accurate general representation of the training set. Because these representations are continuous, multi-dimensional, and  represent the whole training set, we call them `Holographic Representations' and the architectures capable of generating them `Holographic Neural Architectures' (HNAs).

In this work, we introduce the general framework of HNAs as well as an implementation of the concept. We tested our models on several datasets:  MNIST \citep{mnist}, Fashion-MNIST \citep{fashionmnist}, SVHN \citep{svhn}, Olivetti \citep{olivetti}, TCGA \citep{tcga} and IEDB \citep{iedb}. We show that HNAs are inherently highly resistant to input noise. We report that HNAs can be used to build noise resistant generative networks even on very small datasets, as well as state-of-the-art regression models. Finally, we also explore the effects of different activation functions on the performances of HNAs. Our results show that sine types activations outperform more widely used non-linearities.

%% file: arch.tex
\section{Architecture}

\begin{figure}[h]
\centering
\includegraphics[width=0.5\textwidth]{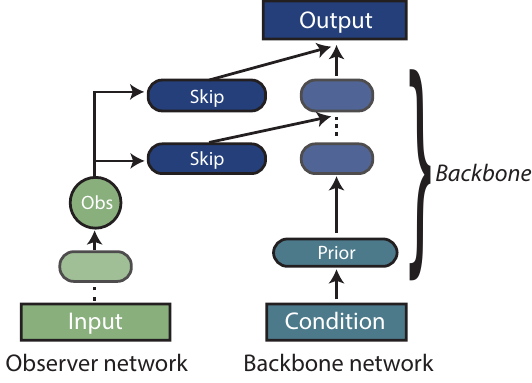}
\caption{The HNA used in this work}
\label{fig:architecture}
\end{figure}

We implemented HNAs as two joined networks as depicted in figure~\ref{fig:architecture}: a backbone network, that learns distributions over the training data, and an observer network whose output  modulates the activations of the backbone network neurons. The backbone is itself composed of two parts: an optimized prior, and a posterior
network that integrates the prior with the information received from the observer. Only the observer network receives information specific to the desired output. The combination in the backbone of the prior and the specific information encoded by the observer produces an approximation of the desired output. Interestingly, without the measurement made by the observer network, the output of the backbone network alone represents an undifferentiated state, i.e. it is a distribution of all possible outputs. Furthermore, because the output of the observer network is directly connected to all neurons inside the backbone network, theses neurons are entangled. In other words, their activations are synchronized in a way that causes the output of the network to collapse into values representing a more restricted set of possible outputs. Because the role of the backbone network is to explicitly encode the space of shared variations among examples, it allows for the specific information of the observer to be encoded in a very low dimensionality space. For this paper we used a single neuron with a sine activation as the output of the observer network, reducing all the specific information about the input to a single, bounded and continuous dimension.

Figure~\ref{fig:architecture} shows the architecture that we used to implement our HNAs. The prior can be easily conditioned (for example on the class label), by having one distribution for every condition. Here we used Gaussian mixtures as priors and optimized the parameters of each mixture during training. Every hidden layers in the backbone as well as the output layer receive the output of the previous layer, concatenated with a skip connection to the observer's output. These skip connections are expanded to match the size of the previous hidden layer. Therefore, the activation of the layers of the backbone network are computed using the following formula:

\begin{equation}
h_{n} = f(W_{n} \cdot (h_{n-1} | s_{n})) + b_{n}
\end{equation}

Where $h_{n}$ is the activation of the $n^{th}$ layer, $f$ is an activation function, $s_{n}$ is the $n^{th}$ skip connection of the same size as $h_{n-1}$, $W_{n}$ a parameter matrix of size $N \times 2N$, $b_{n}$ is a bias parameter, $|$ indicates a concatenation and $\cdot$ is matrix multiplication. Thanks to the skip connections, the integration of the prior with the observer's measurement is performed throughout the network. Every layer in the backbone network has its activation modified by the output of the observer, both directly and through the activation of its previous layer.

%% file: hgn.tex
\section{Holographic Generative Networks}

Generative networks try to learn the distribution of training examples in order to generate new samples. Because HNAs explicitly learn priors over the training data, they should perform well as generative networks. We designed Holographic Generative Networks (HGN) as auto-encoders. Here the information received by the observer is the image to be reconstructed, while the backbone network receives random samples drawn from Gaussian mixtures conditioned on the image class. To ensure the reproducibility of our results, and assess the ease of training of HGNs, we used a single architecture for all experiments in this section regardless of the dataset: MNIST\citep{mnist}, Fashion-MNIST\citep{fashionmnist}, SVHN\citep{svhn}, Olivetti\citep{olivetti}. All networks minimized a MSE reconstruction loss and were trained using adagrad\citep{duchi2011adaptive} with a learning rate of 0.01. Layers were initialized using Glorot initialization\citep{glorot2010understanding}, Gaussian mixtures parameters were initialized using a [0, 1] bounded uniform distribution, each mixture contained 16 1-dimensional Gaussians. We used a layer size of 128, 2 hidden layers for the observer network and 4 for the backbone. All networks were implemented and trained using  the deep learning framework Mariana\citep{mariana}.

\begin{figure}[h]
\centering
\includegraphics[width=\textwidth]{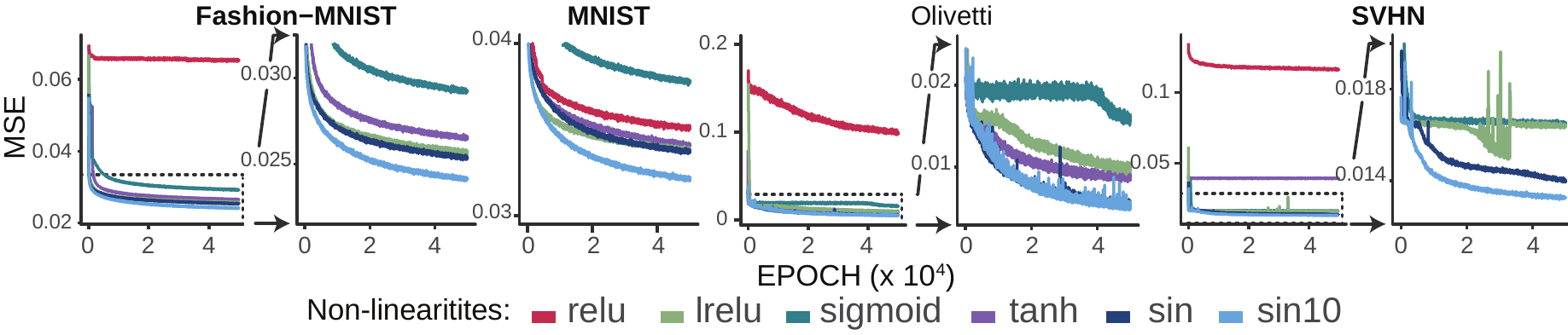}
\caption{Impact of non-linearities on the convergence of HGNs. Sine type activations outperform more widely used non-linearities}
\label{fig:nonlin}
\end{figure}

Because HNAs heavily rely on the modulation of the backbone network by the observer’s output, the backbone representations have to be rich and flexible enough to allow for a wide set of possible modulations. To explore the impact of activation functions on HNAs, we compared convergence performances of networks using the following non-linearities: sigmoid, tanh, relu\citep{relu}, lrelu\citep{lrelu}, and sine. For networks using sine activations we used a normal sine activation for all layers but the last one, for which we used a sine normalized between [0, 1] to match target values. As shown in figure~\ref{fig:nonlin}, HGNs using sine type activation functions consistently converge better than the others. Surprisingly, performances of non-sine activation functions are very dataset dependent. We also found that multiplying the output of the observer by an arbitrary value (here 10) can improve convergence~\ref{fig:nonlin}, we refer to this non-linearity as `sin10'. As shown in figure~\ref{fig:nonlin}, HGNs using sin10 converge better than those using a regular sine function. The only exception being the Olivetti dataset, on which both networks showed very similar performances. These results suggest that HNAs benefit from the non-linear processing performed by sine type functions.

\begin{figure}[h]
\centering
\includegraphics[width=\textwidth]{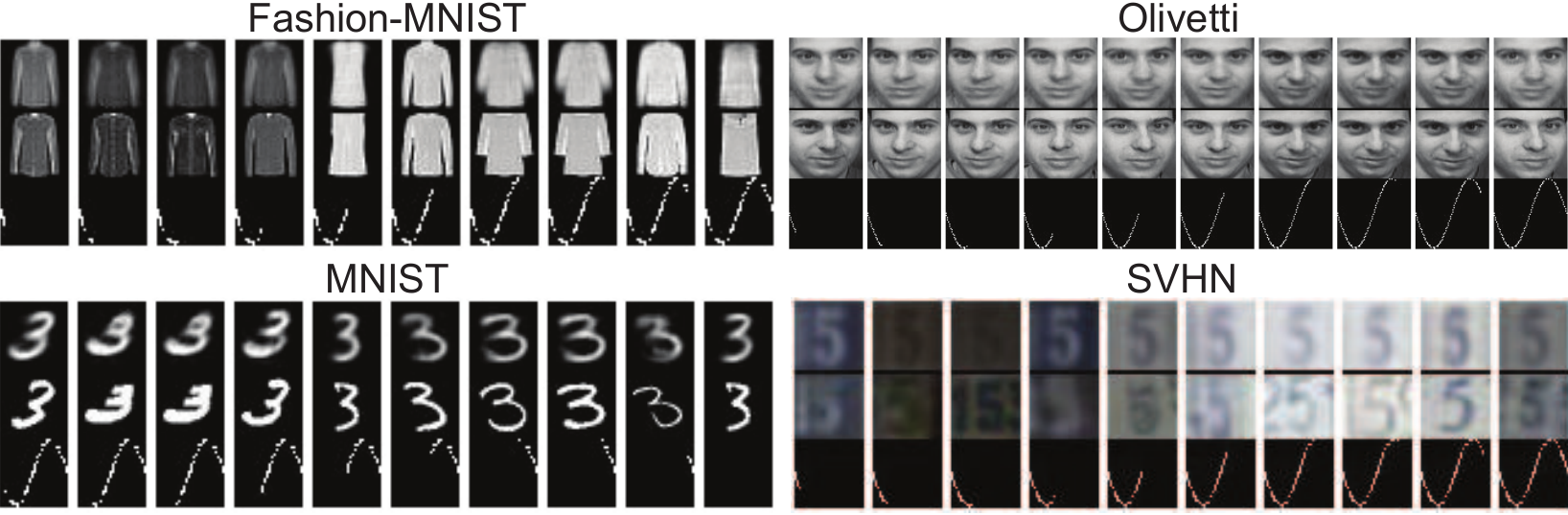}
\caption{HGN generated images. The first row shows generated images using an FSS method. The second row shows the closest images in the training set calculated using an euclidean distance. The third row shows the observer activation.}
\label{fig:recons}
\end{figure}

Sampling from a trained HGN is trivial. After fixing the class label, the whole generation process is controlled by the single output of the observer. Since this value is bounded between [-1, 1], our sampling method simply generates images using an ascending vector of $N$ evenly spaced values between [-1, 1]. We call this method of sampling Full Spectrum Sampling (FSS). It has the advantage of generating samples showing the whole spectrum of what the model can generate with arbitrary precision. Furthermore, because all elements are arranged along a single dimension, we found that examples will generally be arranged in a ``semantic'' way, with similar images corresponding to similar observer activations. This is exemplified in figure~\ref{fig:recons} where we can see the gradual evolution of generated outputs that occur on all datasets.

As shown in figure~\ref{fig:recons}, the same architecture was able to generate realistic images for all datasets. Some of the generated images are however a little blurry. Our hypothesis is that the blur is a consequence of the architecture not having enough capacity to effectively model the distribution of training examples. This is supported by the fact that generations on the Olivetti dataset are much sharper. Of all datasets, the Olivetti dataset is the most challenging for generative models because of the small number of samples. This dataset contains 10 images for every one of the 40 subjects. The small number of examples makes it easy to overfit and harder to generalize. Nonetheless, the HGN was able to generate smooth interpolation between facial expression as shown in figure~\ref{fig:recons}. Using the same architecture as for the other datasets, we were able to generate 100 images for each subject from the 10 available in the training set. Because the model was conditioned on the image class, here the subject, it learned a different movement pattern for each class. These movements include continuous changes of facial expressions and 3D rotations as shown in figure~\ref{fig:recons}. On Olivetti we've found that adding a small fixed amount of stochastic Gaussian noise can slightly improve the smoothness and quality of interpolation. Interestingly, SVHN generations show no side digits. This is due to the lack of correlation between the center digit and the side digits. This behavior is related to the noise reduction performed by HNAs. This aspect is explored further in the next section.

\subsection{Demonstration of Noise Resistance}

An interesting feature of HNAs is their ability to filter out noise from the training set. This generalization property is especially important when the signal to noise ratio cannot be as directly assessed as with images, or when getting noise free examples is virtually impossible.

\begin{figure}[h]
\centering
\includegraphics[width=\textwidth]{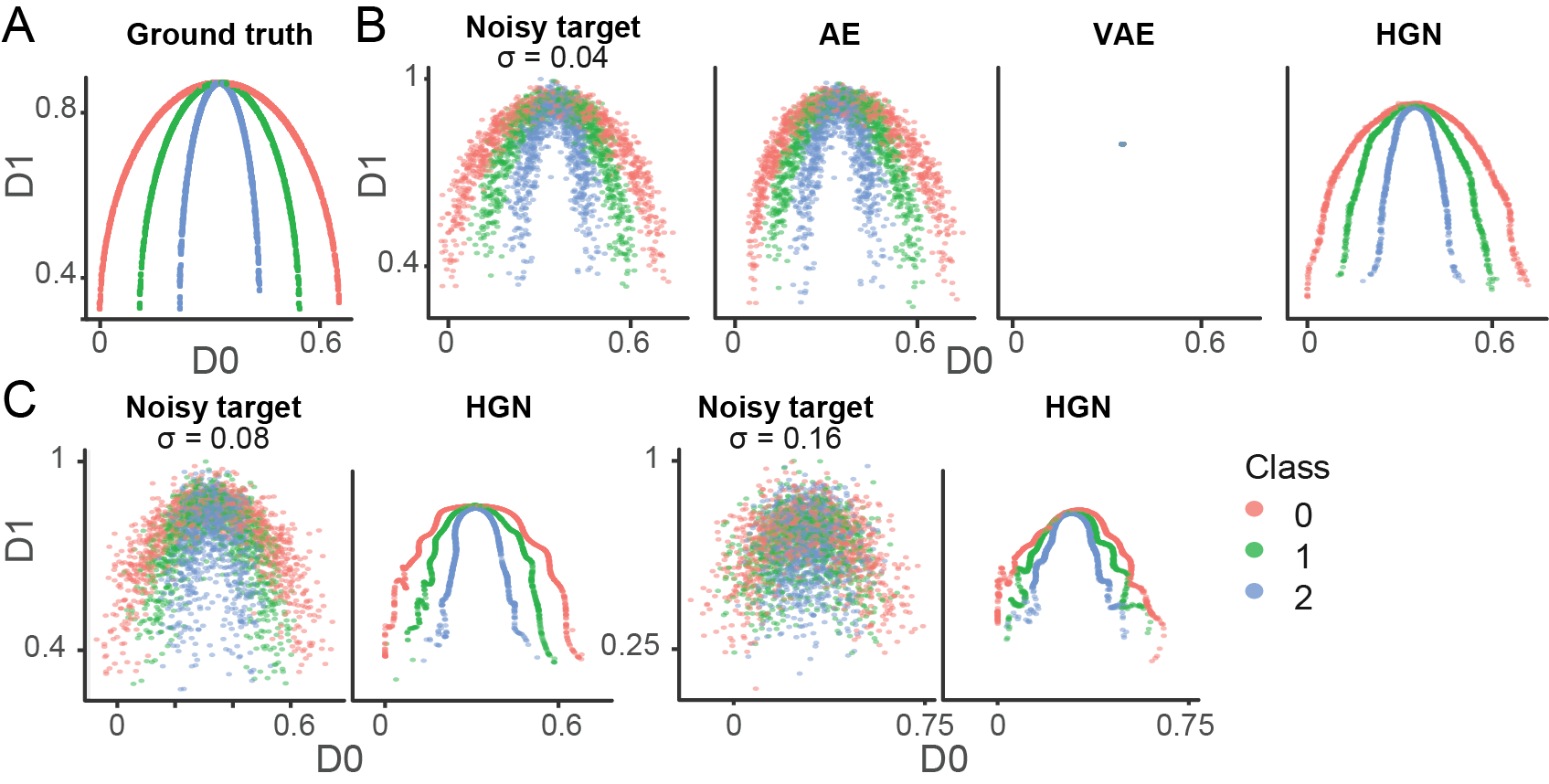}
\caption{Comparison of denoising performances of an HNG, a standard AE, and a VAE tested on a corrupted synthetic toy dataset.}
\label{fig:noiseres1}
\end{figure}

To illustrate the nature of the noise filtering, we generated a synthetic dataset from three concentric crescents, each one corresponding to a single class. For each class, we generated 1000 points and added increasing amounts of normal noise. We then trained an HGN to reproduce the points of each class. Here it is important to note that, contrary to standard denoising Auto-Encoders (dAE) setup where the network receives noisy inputs but the costs are computed on clean targets\citep{dae1,dae2}, here both the input and the target are identical and thus share the same noise. This setup therefore mimics a real life situation where the true target cannot be known. We can see in figure~\ref{fig:noiseres1} that the HGN was able to remove a significant part of the noise in the reconstructions. As expected, the capacity of the network to extract the underlying distributions decreases as we increase the noise standard deviation. However, at high levels of noise (std=0.08), the network is still able to extract clear distinct distributions for each class. At very high levels of noise (std=0.16), the network extracted important features about the original data such as the bell shape and the 3 separate classes. These results show that HGNs are capable of retrieving information about underlying distributions from very corrupted examples. Figure~\ref{fig:noiseres1} also shows the results obtained by a standard AE with relu units and an architecture similar to the backbone network of the HGN, and by a variational autoencoder (VAE). We've used a standard VAE implementation \citep{vae}. The inference model is gaussian with a diagonal covariance, and the generative model is a bernoulli vector:

\begin{align}
z \sim q_{\phi}(z | x) &= \mathcal{N}(\mu_{\phi}(x), \Sigma_{\phi}(x)^2)\\
x \sim p_{\theta}(x | z) &= \mathcal{B}(\pi_{\theta}(z))
\end{align}

We report results with the prior $p(z) = \mathcal{N}(0, \mathtt{I})$ as we have not found that learning the prior during training helps the model perform the denoising operation. Under these conditions, the results from figure~\ref{fig:recons} display a typical failure mode of VAEs when trained on low-dimensionality inputs (as hinted also in \citet{vaefail1}). As expected, the standard AE overfitted the training set and was unable to distinguish the true input signal from the overall image distribution.

We further designed an experiment on the Olivetti and MNIST datasets mimicking the variability encountered in real life data capture. In both cases we corrupted images using the following formula:

\begin{equation}
\widetilde{X} = \mathtt{max}(0, X + \mathcal{N}(0, \Sigma^2))
\end{equation}

The clipping to 0 for negative values simulates a situation where the signal goes below the detection threshold. The corruption was performed only once on the whole dataset, not at every training step. This simulates the real life situation where only a finite number of noisy training examples are available.

\begin{figure}[h]
\centering
\includegraphics[width=\textwidth]{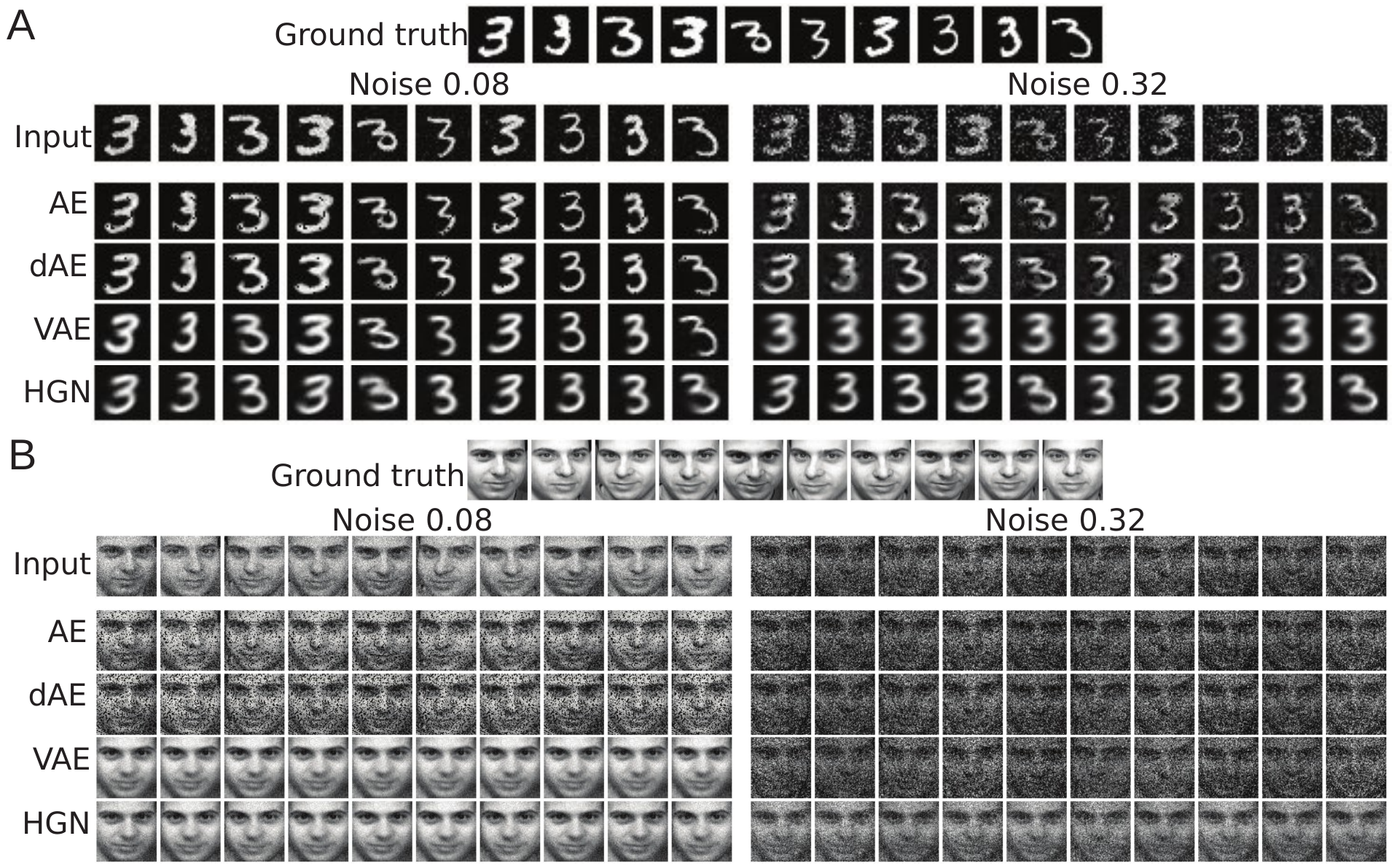}
\caption{Comparison of denoising performances when models are trained on corrupted data of a single class of: (A) MNIST and (B) Olivetti. Reconstructions obtained with a standard AE, a dAE, a VAE and a HGN.}
\label{fig:noise_1c}
\end{figure}

\begin{figure}[h]
\centering
\includegraphics[width=\textwidth]{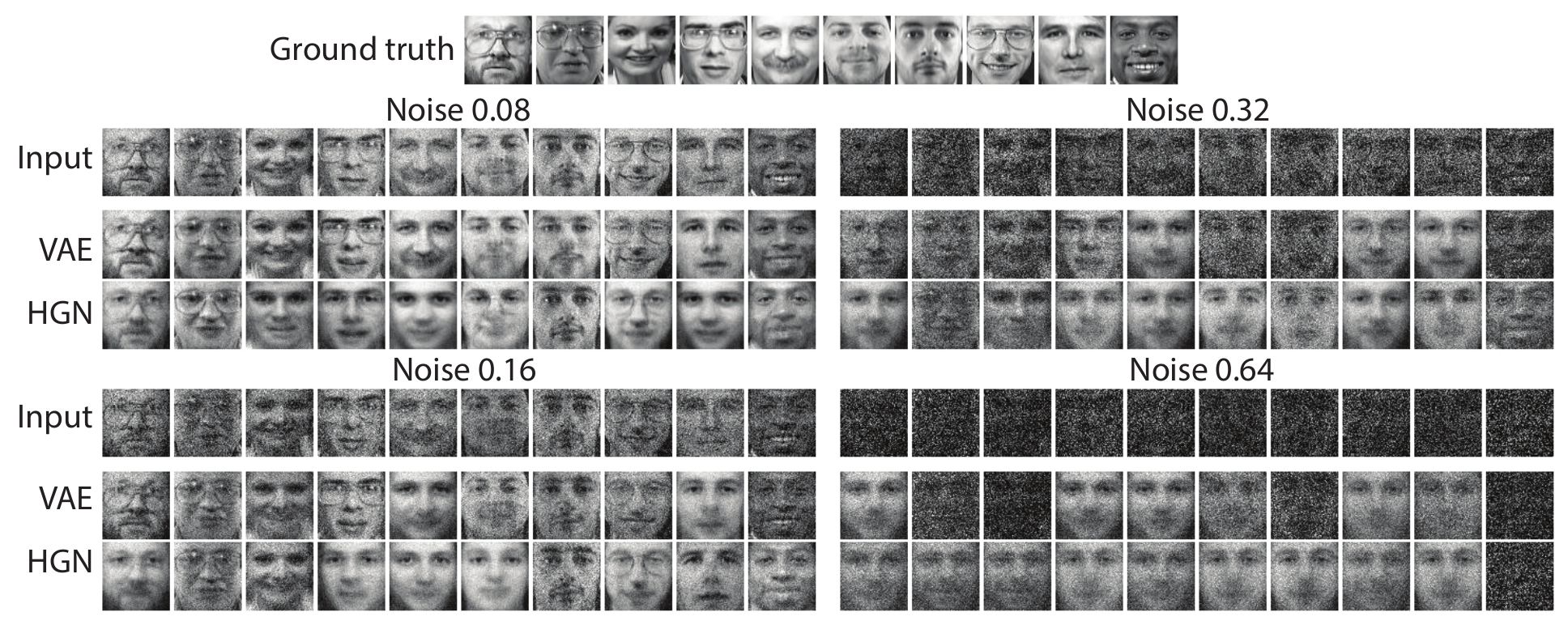}
\caption{Comparison of denoising performances when models are trained on corrupted unlabeled examples. Reconstructions obtained with a VAE and a HGN.}
\label{fig:noise_unlab}
\end{figure}

Using these corrupted datasets we compared the denoising performances of HGN to those of other models. Because standard implementations of AEs, dAEs and VAEs are not conditioned on the class, we trained all models either on a single class for each dataset (figure~\ref{fig:noise_1c}), or on unlabeled datasets by putting all examples under the same class (figure~\ref{fig:noise_unlab}). As shown in figure~\ref{fig:noise_1c}, at low noise levels on MNIST (std=0.08), both the AE and the dAE appear to have overfitted the data. The reconstructions of the VAE and HGN are both noiseless and look similar to each other. On higher noise levels (std=0.32), the dAE was able to remove more of the noise than AE. The VAE output collapsed towards the average image. This shows a failure mode of VAEs. Natural images rarely have a multivariate gaussian structure. Therefore, unless $z$ is lossless, which is not going to be the case in practice, the best-fitting distribution for the generative model will be averaging multiple outputs \citep{vaefail2,elbo}. The HGN reconstructions were not affected by the noise increase.

On the Olivetti dataset, training on a single class means training on 10 images. Figure~\ref{fig:noise_1c} therefore shows the whole training set. At low noise level (std=0.08), the AE and dAE show similar outputs and neither was able to perform any noise reduction. The VAE again collapsed on the average face. The HGN reconstructions however are significantly less noisy than the inputs and also show variability in the face angle. On high noise levels (std=0.32) the HGN was the only model that did not overfit, and was able to remove a major part of the noise.

Figure~\ref{fig:noise_unlab} shows the comparison of a HGN and VAE when trained on an unlabeled, corrupted version of Olivetti. At low noise levels (std=0.08), both models generated faces that are less noisy than the inputs. Interestingly, some of the faces generated by the HGN do not reconstruct the inputs, such as with the face of the third woman from the left. This behavior suggests that the model is relying more on the general representation in the backbone than on the single value received from the observer. The more we increase the noise level, the bigger the difference between the two models. At very high noise levels (std=0.32 and std=0.64) the HGN was more able to extract a face structure than the VAE, and still generate outputs showing some variability. In conclusion, despite being presented with very corrupted inputs from very small datasets, HGNs did not overfit, and were able to filter out a major part of the noise. These results show that HGNs are highly noise resistant.

%% file: bio.tex
\section{Biomedical Datasets}
\subsection{The Cancer Genome Atlas dataset}

Recent years have seen a big increase in the availability of RNA sequencing data \citep{tcga,gtex}. Despite constant improvements in the devices and protocols being introduced, RNA sequencing remains a complex process. Samples are often collected by different experimenters in different conditions. These factors introduce an element of uncertainty that is very difficult to quantify. This renders RNA sequencing, and therefore the gene expressions that are derived from the sequencing, susceptible to noise. Another issue that is inherent to the cost and availability of biological data, is that it can often be impossible to get a large number of RNA sequencing samples for a given condition.

\begin{figure}[h]
\centering
\includegraphics[width=\textwidth]{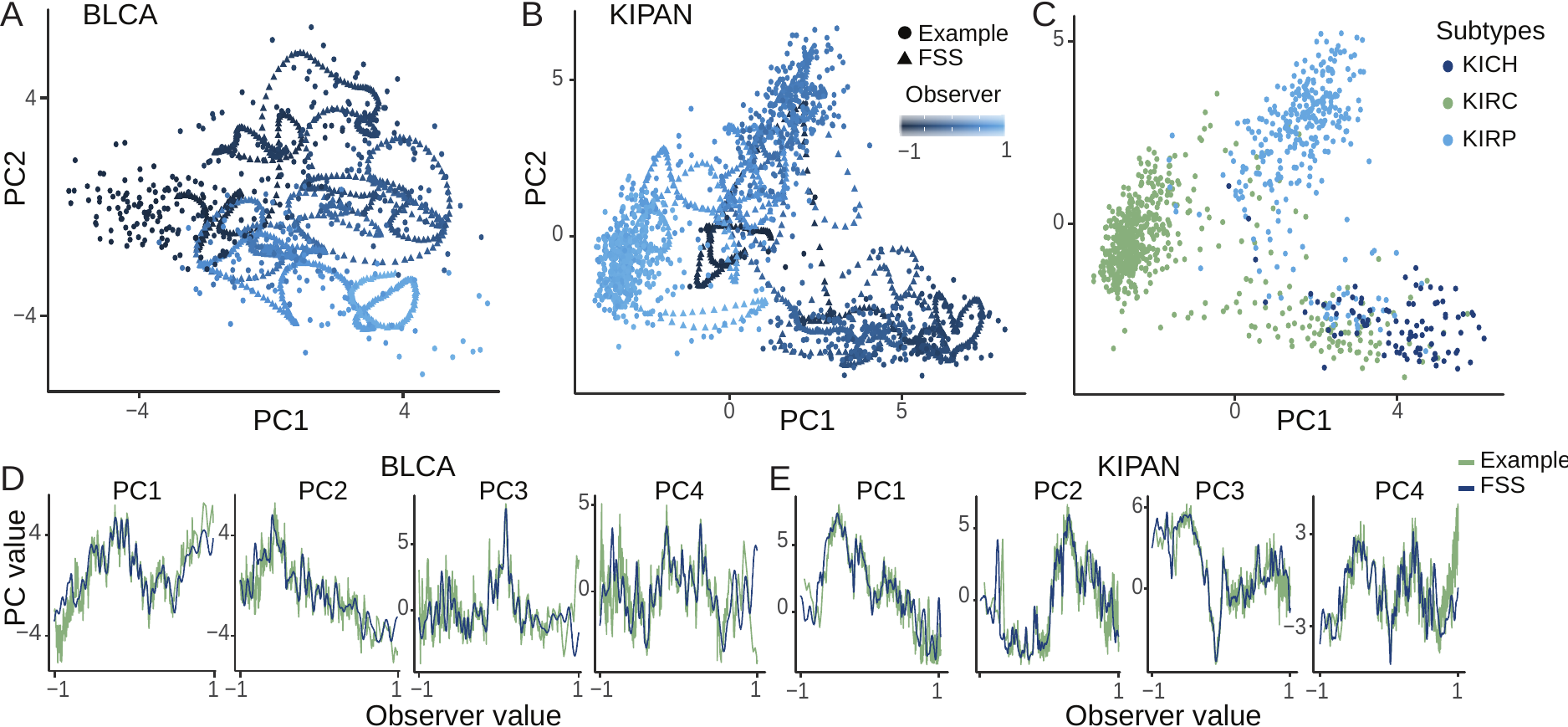}
\caption{PCA of cancer gene expression samples for: (A) bladder cancer samples (BLCA) when training examples and generated samples are reduced using the first 2 principal components of training examples. Round dots represent training examples, triangles represent samples generated using a FSS method. The color represents the activation of the observer. (B) shows the same graph for kidney cancers (KIPAN). (C) Distributions of the 3 cancer subtypes composing KIPAN. The last row shows how well generated samples follow the distribution of training examples when principal components are taken individually for BLCA (D) and KIPAN (E).}
%
%
\label{fig:tcga}
\end{figure}

To investigate how well HGNs can model cancer gene expression data, we trained an HGN on the TCGA dataset \citep{tcga}. This dataset contains \numprint{13246} samples from 37 cancers. The number of samples for each cancer type ranges from 45 to \numprint{1212}. Every sample contains the measured expressions for \numprint{20531} genes. We used the same previous HGN architecture, but with a layer size of 300 neurons and 10 layers in the backbone network. Here the network receives gene expressions instead of images and cancer types instead of image classes. Since gene expressions cannot be plotted and visually evaluated, we elected to reduce the dimensionality of both training examples and reconstructions using a PCA (figure~\ref{fig:tcga}). We chose PCA because it is a linear transformation, therefore any non-linear behavior observed after the dimensionality reduction is due entirely to the network. We computed, for every cancer type, a PCA transformation on the training examples, and used that transformation to reduce both training examples and reconstructions. Reconstructions, therefore, are not only presented from the `point of view' of training examples, but they also have no influence on the PCA computation. The transformation is therefore not biased toward accurately representing reconstructions.

Figure~\ref{fig:tcga} shows how well the distribution of generated gene expressions, models the distribution of training samples when both are reduced to their first 4 principal components. Results on KIPAN are particularly interesting. This cancer type is made of 3 distinct subtypes. Despite the heterogeneity of the samples, the HGN was capable of modeling the 3 subtypes, and of generating intermediate samples between these subtypes. These results strongly suggest that HGNs can accurately model RNA sequencing data. Figure~\ref{fig:tcga} also sheds light on the inner functioning of HGNs as it show the trajectory followed in the input space as the observer value increases. Through the interaction with the backbone network, the 1-dimensional manifold learned by the observer is embedded in the higher-dimensional holographic representation of the backbone. The result is then projected onto the input space.

\subsection{IEDB}

MHC-I associated peptides (MIPs) are small sections of proteins that are presented at the surface of cells by the MHC-I molecules. MIPs present an image of the inner functioning of the cell to the immune system and thus play an essential role in the identification of cancerous and virus-infected cells. Given their central role in immunity, being able to reliably predict the binding affinity of MIPs to MHC-I alleles has been identified as a milestone for the development of virus and cancer vaccines \citep{mhcvac,Backert2015}. Modern high throughput studies routinely generate tens of thousands of MIPs whose binding affinity to the subjects' MHC-I molecules have to be assessed \citep{mhcdset}. Given the impossibility of experimentally measuring such a high number of affinities in the lab, machine learning affinity predictors are now the norm \citep{Backert2015,Granados2014,Granados2012,Laumont2016,Pearson2016}. MHC-I molecules are however encoded by extremely polymorphic genes in the human species ($\sim$\numprint{13324} alleles) \citep{robinson2014ipd}. This makes it impossible to gather tens of thousands of training examples for every single allele and modern predictors suffer from a lack of precision regarding rare alleles. 

We previously showed on the Olivetti dataset that HGNs are capable of learning rich representations from a small number of training examples. Here we tested if this extends in particular to regression models as well. Our Holographic Regression Network (HRN) shares the same architecture as the HGNs. However, the network takes the label of the MHC-I allele instead of the class label, and the sequence of the MIP in amino acids encoded using word embeddings \citep{bengio2003neural} instead of the image. Most MIPs vary between 8 to 11 amino acids in length. To account for this difference, we used a default size of 11 and padded the remaining slots with zeros if needed. The output of the network is the predicted affinity of the MIP sequence to the MHC-I allele. We trained our network on the IEDB database \citep{iedb} that contains \numprint{185157} measurements of affinities of MIPs to MHC-I alleles, for 6 species including humans. We normalized all the measurements between [0, 1] by transforming them using the following formula:

\begin{equation}
X = \log_2 (V+2)
\end{equation}

where $V$ is the vector of measurements, and then by dividing by the maximum value of $X$. IEDB also contains several MIP to MHC-I alleles associations with underspecified binding measurements that only give an upper or lower bound. We elected to keep these examples and use the lower or upper bound as the target value. We evaluated our model on a 80/20 random split of the data. We trained our model on 80\% of the data and tested on the remaining 20\%. 

Our model achieved comparable performances than those reported by the state-of the-art model NetMHC 4.0 \citep{andreatta2015gapped} (see table 1). HRN results show higher Pearson correlation coefficient (PCC) than NetMHC4.0 on all lengths except for the 11-mers. Furthermore, our results also show very good performances on the alleles that were removed in the study \citet{andreatta2015gapped} for having too few examples (less than 20 MIPs). These results show that the ability of HGNs to generalize from very few examples extends to HRNs as well. This suggests that HNAs can improve the performances of regression models, especially when the number of training examples is small. 

\begin{table}[h]
\centering
\begin{tabular}{@{}lll@{}}
\toprule
& NetMHC4.0 & HRN\\
\midrule
8-mer & 0.717 & \textbf{0.761}\\
9-mer & 0.717 & \textbf{0.761}\\
10-mer & 0.744 & \textbf{0.747}\\
11-mer & \textbf{0.706} & 0.691\\
\hline
\hline
All lengths, human MIPs only & - & 0.769\\
All lengths & - & 0.760\\
All lengths, MHC-I alleles with less than 20 ex. & - & 0.953*\\
\bottomrule
\end{tabular}
\caption{Pearson correlation coefficient (PCC) NetMHC4.0 vs HRN. NetMHC4.0 results are taken from \citet{andreatta2015gapped}.\\$*$Training set contains 160 examples, testing 19 examples.}
%
%
\label{tbl:iedb}
\end{table}

%% file: conclusion.tex
\section{Conclusion}

In this work, we have introduced HNAs, a new framework for deep learning capable of deriving holographic representations from training sets. We have shown that HNAs can generalize from very few examples, can be used for both generative models and state-of-the-art regression models, and are highly noise resistant. These characteristics make HNAs particularly well-suited for biological applications where the available training data is limited or noisy. We also showed that, in the context of HNAs, the choice of activation function can dramatically influence convergence. Our results show that sine activations consistently outperforms several more widely used activation functions. Whether these results extend to other type of architectures remains to be investigated. Our experiments also suggest that HNAs are easy to train. All the networks in this work follow the same basic recipe, and we used the exact same architecture for all tasks on image datasets. Finally, a very interesting aspect of HNAs is their ability to project all examples into a single bounded dimension. This strongly suggests that they could also be used as effective dimensionality reduction methods.